\documentclass{article}
\usepackage{spconf,amsmath,graphicx}
\usepackage{multirow}
\usepackage{booktabs}

\title{IMAGE-TO-VIDEO RE-IDENTIFICATION VIA MUTUAL DISCRIMINATIVE KNOWLEDGE TRANSFER}
\name{Pichao Wang, Fan Wang, Hao Li}
\address{Alibaba Group}
\begin{document}
%
\maketitle
\begin{abstract}
The gap in representations between image and video makes Image-to-Video Re-identiﬁcation (I2V Re-ID) challenging, and recent works formulate this problem as a knowledge distillation (KD) process. In this paper, we propose a mutual discriminative knowledge distillation framework to transfer a video-based richer representation to an image based representation more effectively. Specifically, we propose the triplet contrast loss (TCL), a novel loss designed for KD. During the KD process, the TCL loss transfers the local structure, exploits the higher order information, and mitigates the misalignment of the heterogeneous output of teacher and student networks. Compared with other losses for KD, the proposed TCL loss selectively transfers the local discriminative features from teacher to student, making it effective in the ReID. Besides the TCL loss, we adopt mutual learning to regularize both the teacher and student networks training. Extensive experiments demonstrate the effectiveness of our method on the MARS, DukeMTMC-VideoReID and VeRi-776 benchmarks.
\end{abstract}
\section{Introduction}
\label{sec:intro}

Re-identification (Re-ID) targets to retrieve an object among the gallery set that has the same identity with the given query across non-overlapping cameras. Generally speaking, this task involves three main categories: image-based (I2I), video-based (V2V) and Image-to-Video (I2V) Re-ID. Contrary to I2I and V2V Re-ID, I2V Re-ID targets to build bridges between image representations and video representations. Recent works~\cite{gu2019temporal,porrello2020robust} formulate this problem as a knowledge distillation process. Specifically, Gu et al.~\cite{gu2019temporal} transfer temporal information from a video-based teacher network to an image-based student one; Porrello et al.~\cite{porrello2020robust} distill the multiple views knowledge from teacher to student network. In terms of loss function for distillation, in TKP~\cite{gu2019temporal}, feature distance and cross sample distance are used for propagation; in VKD~\cite{porrello2020robust}, logits and pairwise distance are adopted for distillation. All of these distillation losses focus on the global matching, designed for classification tasks~\cite{hinton2015distilling} or representative learning~\cite{tung2019similarity}. However, in the Re-ID task, the discriminative ability of features is more important due to the non-overlapping ID labels of the training and testing set. 

In this paper, inspired by the commonly used triplet loss~\cite{hermans2017defense} for ReID task, we design the triplet contrast loss (TCL) for the knowledge distillation between teacher network and student network. Instead of directly comparing the distance between anchor-positive and anchor-negative pairs, we propose to measure the probabilities of these distances. 
Unlike other losses~\cite{gu2019temporal,porrello2020robust} that match the global 
information between teacher and student, the TCL loss focuses on the local structure consisting of nearest neighbors for each example. Coupled with hard triplet sampling in the student embedding, the discriminative ability of features are enhanced by the soft supervision of the teacher network. The TCL loss mitigates the misalignment of the heterogeneous outputs of teacher and student networks. It also encodes the higher order (third order) structured knowledge in the triplet, and introduces richer information of similarity comparison than the vanilla triplets in the context of knowledge distillation. The TCL loss is complementary to conventional KD losses and can be combined with them to further boost the performance. The TCL improves the performance of I2V Re-ID largely, even without the commonly used cross-entropy loss. Besides the proposed TCL, mutual learning~\cite{zhang2018deep} is adopted in order to regularize both teacher and student learning. Integrating these two techniques, the proposed method is dubbed as mutual discriminative knowledge transfer (MDKT). Extensive experiments demonstrate the effectiveness of the proposed method on MARS~\cite{zheng2016mars}, DukeMTMC-VideoReID~\cite{wu2018exploit} and VeRi-776 benchmarks~\cite{liu2016deep}.

\begin{figure}
\centering
\includegraphics[width=\columnwidth]{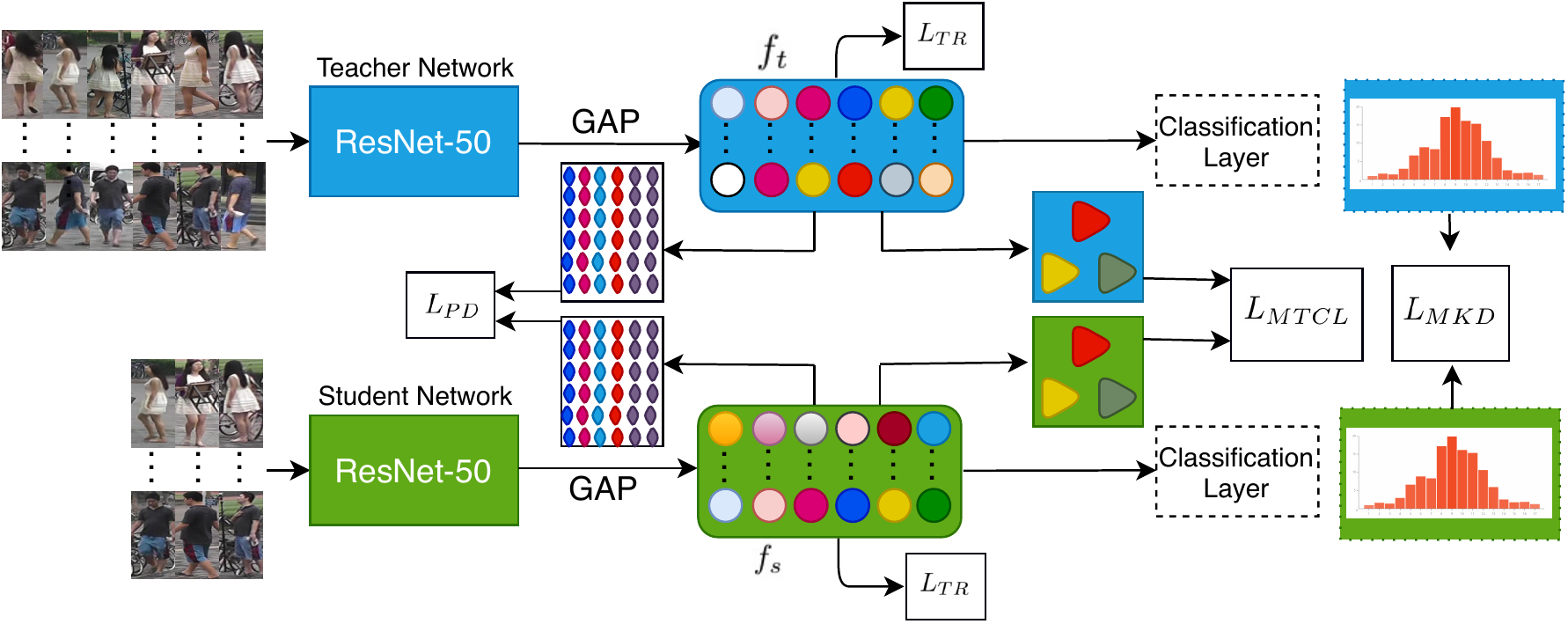}
\caption{The framework of MDKT. }
\label{framework}
\end{figure}

\section{The Proposed Method}\label{proposedmethod}

The framework of MDKT consists of two-stages: 1) the teacher network is trained using the standard V2V Re-ID setting. 2) we feed frames representing different numbers of views as input to the teacher and students networks for view KD using three level distillation losses, as depicted in Figure~\ref{framework}.

\subsection{Teacher Network}
Without loss of generality, ResNet-50~\cite{he2016deep} is used as the backbone network. Following~\cite{gu2019temporal,luo2019bag}, the network is initialized with the weights pretrained on ImageNet. A few amendments as in~\cite{luo2019bag} are included in the architecture.

\textbf{Video Representation.} Given $P$ person video clips $\textit{V} = \{v_{n}\}_{n=1}^{P}$, each $v_{n}$ contains $ T $ frames ($T$ is set to 8 unless otherwise specified). In this paper, the video representation $f_{t}(V_{n})$ for the teacher network is simply computed using the temporal average pooling over all frame features.

\textbf{Teacher Network Optimisation.} The teacher network is trained using the cross-entropy classification loss  and the triplet loss~\cite{hermans2017defense}. The cross-entropy loss $L_{CE}^{t}$ is formulated as:
\begin{equation}\label{lce}
    L_{CE}^{t} = -\boldsymbol{y}_{t} \log \boldsymbol{\hat{y}}_{t}
\end{equation}
where  $t$ represents the teacher model, $\boldsymbol{y}_{t}$ and $\boldsymbol{\hat{y}}_{t}$ represent the one-hot labels and the output of the softmax respectively. The triplet loss $L_{TR}^{t}$ is normally trained on a series of triplets $\{x_{a},x_{p},x_{n} \}$, where $x_{a}$ and $x_{p}$ are videos of the same person, denoted as anchor and positive samples, and $x_{n}$ is from a different person, denoted as the negative sample. The triplet loss is designed to keep $x_{a}$ closer to $x_{p}$ than $x_{n}$ and it is defined as:
\begin{equation}\label{tripletloss}
    L_{TR}^{t}=\sum_{a,p,n}^{N}[\Vert f_{t}(x_{a})-f_{t}(x_{p})\Vert_{2}^{2}-\Vert f_{t}(x_{a}) - f_{t}(x_{n})\Vert_{2}^{2} + \alpha]_{+}
\end{equation}
where $[z]_{+} = max(z,0)$, and $f_{t}(x_{a}),f_{t}(x_{p}),f_{t}(x_{n})$ represent features of the three videos from the teacher network. $\alpha$ is a margin that is enforced between positive and negative pairs. 

\subsection{Mutual Discriminative Knowledge Transfer}
After training the teacher network, following VKD~\cite{porrello2020robust}, we transfer the knowledge lying in multiple views in a teacher-student distillation fashion. We allow the teacher to access frames $I_{t} = (i_{1},i_{2},\cdots, i_{N})$ from different viewpoints  and feed the student with subset of teacher's inputs $I_{s} = (i_{1},i_{2},\cdots, i_{M})$, where the cardinality N \textgreater M (in the experiments, N = 8 and M = 2). The MDKT is formulated as an optimization problem using three level distillation losses:

\textbf{Mutual Logits Distillation.} Similar to VKD~\cite{porrello2020robust}, the logits based knowledge distillation loss~\cite{hinton2015distilling} is adopted to distill the multiple view knowledge from teacher to student:
\begin{equation}\label{ldl}
    L_{KD_{t2s}} = \tau_{1}^{2}KL(y_{t} \Vert y_{s})
\end{equation}
where  $s$ represents the student model; $KL$ denotes the Kullback-Leibler divergence; $y_{t} = softmax(f_{t}(x)/\tau_{1})$ and  $y_{s} = softmax(f_{s}(x)/\tau_{1})$ are the output distributions of teacher and student networks; $\tau_{1}$ is a non-negative temperature, the large the value of $\tau_{1}$, the smoother the output. $L_{KD_{t2s}}$ is a first-order distillation loss and it makes the student keep its predictions consistent with the teacher. Unlike VKD~\cite{porrello2020robust}, another loss from student to teacher for logits distillation is added as follows: 
\begin{equation}
    L_{KD_{s2t}} = \tau_{1}^{2}KL(y_{s} \Vert y_{t})
\end{equation}
The final mutual logits distillation loss is:
\begin{equation}\label{mld}
    L_{MKD} = L_{KD_{t2s}} + L_{KD_{s2t}}
\end{equation}

\textbf{Pairwise Distance in Embedding.} In addition to logits distillation, we also adopt an embedding-based knowledge distillation loss $L_{PD}$~\cite{porrello2020robust}:
\begin{equation}
    L_{PD} = \sum_{(i,j) \in (\substack{B \\ 2})} (D_{t}[i,j] - D_{s}[i,j])^{2}
\end{equation}
where $D_{t}[i,j]=D(f_{t}(I_{t}[i]), f_{t}(I_{t}[j]))$, indicating the distance induced by the teacher between the $i$-th and $j$-th inputs (the same notation $D_{s}[i,j]$ also hold for the student); $B$ denotes the batch size. The $L_{PD}$ loss is a second order loss which encourages the student to mirror the pairwise distances spanned by the teacher.

 \begin{table*}[ht]
\centering
\begin{tabular}{c|c|c|c|c|c|c|c|c|c|c|c|c} 
\hline
\multirow{3}{*}{Models~~} & \multicolumn{4}{c|}{Losses}                                                                                                                                                       & \multicolumn{4}{c|}{MARS}                                                                                   & \multicolumn{4}{c}{Duke-video}                                                                                    \\ 
\cline{2-13}
                          & \multicolumn{1}{l|}{\multirow{2}{*}{$L_{TR}$}} & \multicolumn{1}{l|}{\multirow{2}{*}{$L_{MKD}$}} & \multicolumn{1}{l|}{\multirow{2}{*}{$L_{PD}$}} & \multicolumn{1}{l|}{\multirow{2}{*}{$L_{MTCL}$}} & \multicolumn{2}{c|}{I2V}                             & \multicolumn{2}{c|}{V2V}                             & \multicolumn{2}{c|}{I2V}                             & \multicolumn{2}{c}{V2V}                              \\ 
\cline{6-13}
                          & \multicolumn{1}{l|}{}                     & \multicolumn{1}{l|}{}                      & \multicolumn{1}{l|}{}                     & \multicolumn{1}{l|}{}                        & \multicolumn{1}{l|}{cmc1} & \multicolumn{1}{l|}{mAP} & \multicolumn{1}{l|}{cmc1} & \multicolumn{1}{l|}{mAP} & \multicolumn{1}{l|}{cmc1} & \multicolumn{1}{l|}{mAP} & \multicolumn{1}{l|}{cmc1} & \multicolumn{1}{l}{mAP}  \\ 
\hline
TR                  & $\surd$                               & $\times$                                          & $\times$                                         & $\times$                                            & 76.77                     & 66.85                    & 84.55                     & 74.23                    & 78.24                     & 70.66                    & 88.24                     & 84.96                     \\ 

\hline
TCL                 & $\times$                                         & $\times$                                          & $\times$                                         & $\surd$                                             & 80.71                     & 71.56                    & 86.82                     & 78.04                    & 82.69                     & 79.26                    & 93.38                     & 92.01                     \\ 
\hline
TR+TCL              &   $\surd$                                        & $\times$                                          & $\times$                                         &     $\surd$                                         & 81.16                     & 72.91                    & 86.36                     & 78.68                    & 83.65                     & 80.32                    & 95.01                     & 93.22                     \\ 
\hline
KD+PD+TCL           & $\times$                                         & $\surd$                                           &     $\surd$                                      &   $\surd$                                           & 84.70                     & 77.56                    & 89.19                     & 82.53                    & 86.32                     & 84.57                    & \textbf{95.58}                     & 93.94                     \\ 
\hline
TR+KD+PD            &   $\surd$                                        &   $\surd$                                         &    $\surd$                                       & $\times$                                            & 83.96                     & 77.43                    & 88.89                     & 82.47                    & 85.04                     & 83.97                    & 95.01                     & 93.69                     \\ 
\hline
TR+KD+TCL           &       $\surd$                                    &     $\surd$                                       &  $\times$                                         &     $\surd$                                         & 83.59                     & 76.28                    & 88.69                     & 81.83                    & 84.90                     & 83.89                    & 94.87                     & 93.56                     \\ 
\hline
TR+PD+TCL           &    $\surd$                                       &  $\times$                                          &     $\surd$                                      &  $\surd$                                            & 85.33                     & 77.90                    & 89.01                     & 82.65                    & 84.90                     & 83.74                    & 95.30            & \textbf{93.94}            \\ 
\hline
ALL                 &           $\surd$                                &  $\surd$                                          &     $\surd$                                      &   $\surd$                                           & \textbf{85.65}            & \textbf{78.02}           & \textbf{89.48}            & \textbf{82.90}           & \textbf{86.78}            & \textbf{84.82}           & 95.26                     & 93.83                     \\
\hline
\end{tabular}
\caption{Ablation study on the impact of loss terms on MARS and Duke-video datasets using ResNet-50.}\label{table3}
\end{table*}


\textbf{Triplet Contrast Loss for Discriminative Transfer.} The two abovementioned distillation losses mainly target to address the representation learning (global matching), but neglect the transfer of discrimination ability (local structure). For Re-ID tasks, the disrcriminative feature learning is more important as the labels between the training set and testing set are different. In order to address this in the context of I2V Re-ID, we propose a third order distillation loss, the triplet contrast loss (TCL), inspired by the vanilla triplet loss. 

In vanilla triplet loss (Eq.~\ref{tripletloss}), the distances between anchor-positive and anchor-negative pairs are used for discriminative feature learning. However, in knowledge distillation, the outputs of two networks are heterogeneous, the absolute distances between these pairs are not well aligned. In order to mitigate the misalignment, we propose to measure the probability of the two distances. Let $d_{a2p}^{t} = \Vert f_{t}(x_{a})-f_{t}(x_{p}) \Vert_{2}^{2}$ and $d_{a2n}^{t} = \Vert f_{t}(x_{a})-f_{t}(x_{n})\Vert_{2}^{2}$, and the probability is defined as:
\begin{equation}\label{papn}
    p_{apn_{\tau_{2}}} = \frac{\exp(-d_{a2p}^{t}/\tau_{2})}{\exp(-d_{a2p}^{t}/\tau_{2}) + \exp(-d_{a2n}^{t}/\tau_{2})}
\end{equation}

 $p_{apn}$ measures how much the anchor is closer to the positive than the negative. If the anchor is closer to the positive than the negative,  $p_{apn}$ is large, otherwise it is small. Compared with the vanilla triplet which accumulates the absolute differences, the $p_{apn}$ brings higher order (third order) similarity comparison, encoding the structural information in the triplet. To transfer the local structure, batch hard sample mining~\cite{hermans2017defense} is adopted to make the local structure consist of nearest neighbors for each example.

\begin{table}
\centering
\begin{tabular}{|c|l|l|l|l|} 
\hline
\multirow{3}{*}{Models}              & \multicolumn{4}{c|}{MARS}                                                                                                                                  \\ 
\cline{2-5}
                                     & \multicolumn{2}{c|}{I2V}                                                    & \multicolumn{2}{c|}{V2V}                                                     \\ 
\cline{2-5}
                                     & cmc1                                 & mAP                                  & cmc1                                 & mAP                                   \\ 
\hline
freeze teacher                       & \multicolumn{1}{c|}{85.10}           & \multicolumn{1}{c|}{77.65}           & \multicolumn{1}{c|}{89.44}           & \multicolumn{1}{c|}{82.79}            \\ 
\hline
without mutual                           & \multicolumn{1}{c|}{85.33}           & \multicolumn{1}{c|}{77.77}           & \multicolumn{1}{c|}{89.22}           & \multicolumn{1}{c|}{82.80}            \\ 
\hline
with mutual                          & \multicolumn{1}{c|}{\textbf{85.65} } & \multicolumn{1}{c|}{\textbf{78.02} } & \multicolumn{1}{c|}{\textbf{89.48} } & \multicolumn{1}{c|}{\textbf{82.90} }  \\ 
\hline
\multicolumn{1}{|l|}{}               & \multicolumn{4}{c|}{Duke-video}                                                                                                                                  \\ 
\hline
\multicolumn{1}{|l|}{}               & \multicolumn{2}{c|}{I2V}                                                    & \multicolumn{2}{c|}{V2V}                                                     \\ 
\hline
\multicolumn{1}{|l|}{}               & \multicolumn{1}{c|}{cmc1}            & \multicolumn{1}{c|}{mAP}             & \multicolumn{1}{c|}{cmc1}            & \multicolumn{1}{c|}{mAP}              \\ 
\hline
\multicolumn{1}{|l|}{freeze teacher} & 86.65                                & 84.58                                & 95.09                                & 93.70                                 \\ 
\hline
\multicolumn{1}{|l|}{without mutual}     & 86.63                                & 84.72                                & 95.10                                & 93.51                                 \\ 
\hline
\multicolumn{1}{|l|}{with mutual}    & \textbf{86.78}                       & \textbf{84.82}                       & \textbf{95.26}                       & \textbf{93.83}                        \\
\hline
\end{tabular}
\caption{Ablation study on the impact of mutual learning.}\label{table4}
\end{table}


With the $p_{apn_{\tau_{2}}}^{t}$ from teacher and $p_{apn_{\tau_{2}}}^{s}$ from student, we define the distribution $P_{apn_{\tau_{2}}}^{t} = [p_{apn_{\tau_{2}}}^{t}, 1-p_{apn_{\tau_{2}}}^{t}]$ and $P_{apn_{\tau_{2}}}^{s} = [p_{apn_{\tau_{2}}}^{s}, 1-p_{apn_{\tau_{2}}}^{s}]$. Thus, the TCL loss between teacher and student is formulated as:
\begin{equation}
    L_{TCL_{t2s}}=\sum_{a,p,n}^{N}KL(P_{apn_{\tau_{2}}}^{t} \Vert P_{apn_{\tau_{2}}}^{s})
\end{equation}
    
Similar to the mutual logits distillation, we also adopt the matching between student and teacher for TCL loss:
\begin{equation}
    L_{TCL_{s2t}}=\sum_{a,p,n}^{N}KL(P_{apn_{\tau_{2}}}^{s} \Vert P_{apn_{\tau_{2}}}^{t})
\end{equation}

Our final mutual TCL loss for final optimization is defined as:
\begin{equation}\label{mtcl}
     L_{MTCL} =  L_{TCL_{t2s}} +  L_{TCL_{s2t}}
\end{equation}


\subsection{The Objective Function}
Besides the three level distillation losses, we also adopt the widely used triplet loss for both teacher and student networks, formulated as:
\begin{equation}
    L_{TR} = L_{TR}^{t} + L_{TR}^{s}
\end{equation}
Different from TKP~\cite{gu2019temporal} and VKD~\cite{porrello2020robust}, we discard the cross-entropy classification loss, as we find that our proposed MDKT can better learn discriminative features even without the classification loss. The final objective function is formulated as the combination of the four losses:
\begin{equation}\label{finalL}
    L = L_{TR} + \alpha L_{MKD} + \beta L_{PD} + \gamma L_{MTCL}
\end{equation}


\subsection{I2V Re-ID Testing}
In the test stage, each query is a static image and the gallery set consists of video clips.  In our work, one single network is used for both image and video feature extraction. Herein, we treat the still image as a video. After feature extraction, the distances between the query feature and each gallery video features are calculated and the I2V retrieval is conducted according to the distances.

\section{Experiments}
\label{sec:experiments}
\subsection{Datasets and Implementation Details}
We evaluate our method on MARS~\cite{zheng2016mars}, DukeMTMC-VideoReID (Duke-video)~\cite{wu2018exploit} and VeRi-776 (VeRi)~\cite{liu2016deep} benchmarks. 
The same training hyper-parameters in VKD~\cite{porrello2020robust} are adopted in this paper. We keep the $\tau_{1}$ = 10 (Equation~\ref{ldl}), $\tau_{2}$ = 4 (Equation~\ref{papn}), $\alpha$ = $10^{-1}$ (Equation~\ref{finalL}), $\beta$ = $10^{-4}$ (Equation~\ref{finalL}) and $\gamma$ = 1000 (Equation~\ref{finalL}) in all experiments. The teacher is set to training mode during distillation.

 \begin{table}
\centering
\begin{tabular}{|c|l|l|l|l|} 
\hline
\multirow{3}{*}{Models}               & \multicolumn{4}{c|}{MARS}                                                                                                                                  \\ 
\cline{2-5}
                                      & \multicolumn{2}{c|}{I2V}                                                    & \multicolumn{2}{c|}{V2V}                                                     \\ 
\cline{2-5}
                                      & cmc1                                 & mAP                                  & cmc1                                 & mAP                                   \\ 
\hline
with CE loss                          & \multicolumn{1}{c|}{84.29}           & \multicolumn{1}{c|}{76.94}           & \multicolumn{1}{c|}{88.69}           & \multicolumn{1}{c|}{82.23}            \\ 
\hline
without CE loss                       & \multicolumn{1}{c|}{\textbf{85.65} } & \multicolumn{1}{c|}{\textbf{78.02} } & \multicolumn{1}{c|}{\textbf{89.48} } & \multicolumn{1}{c|}{\textbf{82.90} }  \\ 
\hline
\multicolumn{1}{|l|}{}                & \multicolumn{4}{c|}{Duke-video}                                                                                                                                  \\ 
\hline
\multicolumn{1}{|l|}{}                & \multicolumn{2}{c|}{I2V}                                                    & \multicolumn{2}{c|}{V2V}                                                     \\ 
\hline
\multicolumn{1}{|l|}{}                & \multicolumn{1}{c|}{cmc1}            & \multicolumn{1}{c|}{mAP}             & \multicolumn{1}{c|}{cmc1}            & \multicolumn{1}{c|}{mAP}              \\ 
\hline
\multicolumn{1}{|l|}{with CE loss}    & 85.83                                & 84.66                                & 94.07                                & 93.44                                 \\ 
\hline
\multicolumn{1}{|l|}{without CE loss} & \textbf{86.78}                       & \textbf{84.82}                       & \textbf{95.26}                       & \textbf{93.83}                        \\
\hline
\end{tabular}
\caption{Ablation study on the impact of cross-entropy loss.} \label{table5}
\end{table}

 \begin{table} [t]
\begin{center}
\begin{tabular}{c|ccc}
\toprule
Method & top1 & top5 & mAP \\
\midrule
P2SNet~\cite{wang2017p2snet} & 55.3 & 72.9 & -   \\
TMSL~\cite{zhang2017image} & 56.5 & 70.6 & -   \\
TKP~\cite{gu2019temporal} & 75.6 & 87.6 & 65.1  \\
STE-NVAN~\cite{liu2019spatially} & 80.3 & - & 68.8  \\
NVAN~\cite{liu2019spatially} & 80.1 & - & 70.2  \\
MGAT~\cite{bao2019masked} & 81.1 & 92.2 & 71.8  \\
READ~\cite{shimread} & 81.5 & 92.1 & 70.4 \\
VKD~\cite{porrello2020robust} & 83.9 & 93.2 & 77.3 \\
\midrule
MDKT & \textbf{85.7} & \textbf{93.3} & \textbf{78.0} \\
\bottomrule
\end{tabular}
\caption{Comparison with SOTA methods on MARS dataset.}
\label{table6}
\end{center}
\vspace{-5mm}
\end{table}


 \subsection{Impact of Loss Terms}
 
We perform a thorough ablation study on the final loss terms (Equation~\ref{finalL}) on MARS and Duke-video datasets, and the results are shown in Table~\ref{table3}. It can be seen that results are largely degraded only using triplet loss (TR model) for training. The reason is that with fewer views input, the batch hard sample mining could not pick up rich triplets for training. Compared with the TR model, the TCL model shows greatly improved performance by using only the mutual triplet contrast loss. The three level distillation losses play a crucial role for performance gain (KD+PD+TCL model). The distillation losses coupled with triplet loss achieves the best results for I2V Re-ID setting. 

 \subsection{Impact of Mutual Learning}
In our proposed MDKT, we adopt mutual learning to regularize both the teacher and student training. In this section, we investigate the role of mutual learning. Three models are designed, namely, freezing the teacher parameters, without mutual learning in Equation~\ref{mld} and ~\ref{mtcl} (without student to teacher loss term), and with mutual learning. The results are illustrated in Table~\ref{table4}. From this Table it can be seen that mutual learning can improve the performances both for I2V and V2V settings.

 \begin{table} [t]
\begin{center}
\begin{tabular}{c|ccc}
\toprule
Method & top1 & top5 & mAP \\
\midrule
STE-NVAN~\cite{liu2019spatially} & 42.2 & - & 41.3  \\
TKP~\cite{gu2019temporal} & 77.9 & - & 75.9  \\
NVAN~\cite{liu2019spatially} & 78.4 & - & 76.7  \\
VKD~\cite{porrello2020robust} & 85.6 & 93.9 & 83.8 \\
READ~\cite{shimread} & 86.3 & 94.4 & 83.4 \\
\midrule
MDKT & \textbf{86.8} & \textbf{94.9} & \textbf{84.8} \\
\bottomrule
\end{tabular}
\caption{Comparison with SOTA methods on Duke-video dataset.}
\label{table7}
\end{center}
\vspace{-5mm}
\end{table}


 \begin{table} [t]
\begin{center}
\begin{tabular}{c|ccc}
\toprule
Method & top1 & top5 & mAP \\
\midrule
PROVID~\cite{liu2017provid} & 76.8 & 91.4 & 48.5  \\
VFL-LSTM~\cite{alfasly2019variational} & 88.0 & 94.6 & 59.2  \\
RAM~\cite{liu2018ram} & 88.6 & - & 61.5  \\
VANet~\cite{chu2019vehicle} & 89.8 & 96.0 & 66.3  \\
PAMTRI~\cite{tang2019pamtri} & 92.9 & 92.9 & 71.9  \\
SAN~\cite{qian2020stripe} & 93.3 & 97.1 & 72.5  \\
VKD~\cite{porrello2020robust} & 95.2 & 98.0 & 82.2 \\
\midrule
MDKT & \textbf{96.0} & \textbf{99.3} & \textbf{83.4} \\
\bottomrule
\end{tabular}
\caption{Comparison with SOTA methods on VeRi dataset.}
\label{table8}
\end{center}
\vspace{-5mm}
\end{table}

 \subsection{Impact of Cross-Entropy Loss}
 Unlike TKP~\cite{gu2019temporal} and VKD~\cite{porrello2020robust}, in our final objective function (Equation~\ref{finalL}), we do not adopt the commonly used cross-entropy classification loss. We find that by introducing the TCL loss, coupled with other three loss terms, the network can learn discriminative features for Re-ID. After adding the cross-entropy classification loss, the final performance will be negatively affected. The results are shown in Table~\ref{table5}.

 \subsection{Comparison with State-of-the-art Methods}
 
 The proposed approach is compared with state-of-the-art I2V Re-ID methods on the MARS, Duke-video and VeRi datasets. The results are presented in Table~\ref{table6}, Table~\ref{table7} and Table~\ref{table8}, respectively.  From these three Tables, we can see that our method clearly outperforms other competitors, almost on all metrics including top-1, top-5, and mAP. Specifically, in terms of top-1 and mAP, our method outperforms by 1.8\% and 0.7\%  on MARS, and 0.5\% and 1.4\% on Duke-video. On the VeRi dataset, our method gains 1.2\% boost on mAP.

\section{Conclusion}
\label{sec:conclusion}
In this paper, we propose a mutual discriminative knowledge transfer method for I2V ReID. The proposed method takes advantage of triplet for local discriminative feature learning and aligns the heterogeneous outputs of teacher and student networks. Coupled with the mutual learning, the proposed method achieves state-of-the-art results on three datasets, covering person and vehicle re-identification.

\bibliographystyle{IEEEbib.bst}
\bibliography{icassp}

\end{document}